\documentclass[a4paper, 12pt]{article}
\usepackage[utf8]{inputenc}

\usepackage[T1]{fontenc}
\usepackage[english]{babel}
\usepackage{caption}
\usepackage{booktabs}
\usepackage{threeparttable} 
\usepackage{enumitem}

\usepackage{graphicx} 
\usepackage{lscape} 
\usepackage[margin=1in]{geometry} 
\usepackage{etoolbox}

\usepackage{dcolumn} 
\usepackage{setspace} 
\AtBeginEnvironment{verse}{\singlespacing} 
\AtBeginEnvironment{tabular}{\singlespacing}
\AtBeginEnvironment{table}{\singlespacing}

\usepackage{caption} 
\usepackage{amsmath,amssymb} 

\usepackage[dvipsnames]{xcolor} 
\newcommand\myshade{85} 
\colorlet{mylinkcolor}{violet}
\colorlet{mycitecolor}{YellowOrange}
\colorlet{myurlcolor}{Aquamarine}

\usepackage{hyperref}
\hypersetup{
  linkcolor  = mylinkcolor!\myshade!black,
  citecolor  = mycitecolor!\myshade!black,
  urlcolor   = myurlcolor!\myshade!black,
  colorlinks = true,
}
\usepackage{apacite} 
\usepackage{natbib}

\usepackage{tcolorbox}
\usepackage{listings}

\hypersetup{
pdftitle={Concept Navigation and Classification via Open-Source Large Language Model Processing},
pdfauthor={Mael D. Kubli},
pdfkeywords={Artificial intelligence, Information Extraction, Quantitative Text Analysis, Machine Learning, Latent Construct Classification, Large Language Model},
}

\title{Concept Navigation and Classification via Open-Source Large Language Model Processing}

\date{\today}

\author{ \href{https://orcid.org/0000-0002-5592-9648}{\includegraphics[scale=0.06]{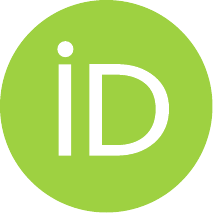}\hspace{1mm}Maël D. Kubli}\thanks{https://www.maelkubli.ch} \\
	Department of Political Science\\
	University of Zurich\\
	\texttt{kubli@ipz.uzh.ch} \\
}

\providecommand{\keywords}[1]
{
  \small	
  \textbf{\textit{Keywords:}} Artificial intelligence, \and Information Extraction, \and Quantitative Text Analysis, \and Machine Learning, \and Latent Construct Extraction, \and Large Language Model
}

\begin{document}
\bibliographystyle{apacite}

\maketitle
\begin{abstract}
\singlespacing

This paper presents a novel methodological framework for detecting and classifying latent constructs, including frames, narratives, and topics, from textual data using Open-Source Large Language Models (LLMs). The proposed hybrid approach combines automated summarization with human-in-the-loop validation to enhance the accuracy and interpretability of construct identification. By employing iterative sampling coupled with expert refinement, the framework guarantees methodological robustness and ensures conceptual precision. Applied to diverse data sets, including AI policy debates, newspaper articles on encryption, and the 20 Newsgroups data set, this approach demonstrates its versatility in systematically analyzing complex political discourses, media framing, and topic classification tasks.

\hspace{0.5cm}


\end{abstract}
\newpage

\section{Introduction}
The advent of Large Language Models (LLMs) has significantly altered the landscape of natural language processing (NLP) and computational linguistics. These models, trained on extensive data sets, exhibit remarkable capabilities in understanding and generating human-like text. LLMs such as GPT-4, Gemini, and LLaMA have demonstrated proficiency in various tasks such as summarization, translation, and text classification, fundamentally changing the way textual data are processed and analyzed \citep{tornberg2023chatgpt, gilardi2023chatgpt, alizadeh2023open, lam2024concept}. This transformative potential is particularly relevant for extracting and classifying latent constructs, which are abstract, unobservable concepts inferred from textual data. Examples include frames, which shape how issues are understood; narratives, which structure events or arguments into coherent meaning; and topics, which group semantically related words to reflect recurring themes in a text. These constructs are essential for understanding political and social discourses. They reveal patterns in framing strategies, ideological bias, and issue salience, thereby clarifying how information is structured to influence audiences or shape policy debates. They also uncover patterns in extensive textual corpora—for example, by showing how news outlets frame political debates, identifying recurring narratives in social media discussions, or spotlighting emerging topics in legislative proceedings. However, extracting such information remains a traditionally difficult task laden with challenges in interpretability, adaptability, and scalability.

Identifying and analyzing latent constructs in the social sciences and humanities is critical to understanding complex discourses and narratives. Specifically, focusing on frames, narratives, and topics enables researchers to address key questions in communication and political science, such as how different groups define problems, justify solutions, and influence public opinion. Although traditional quantitative methods such as Latent Dirichlet allocation (LDA), Structural Topic Models (STM), and BERTopic have been widely used, recent research highlights their limitations in reliably capturing nuanced textual semantics and contextual distinctions, especially when analyzing frames and narratives \citep{eisele2023capturing, mu2024large}. In particular, these classical methods often struggle to differentiate subtle interpretative constructs from mere clusters of co-occurring words, resulting in ambiguous and oversimplified outputs \citep{eisele2023capturing}. In contrast, emerging studies show that Large Language Models (LLMs) significantly outperform these traditional approaches, offering richer semantic understanding, better contextual accuracy, and improved coherence in both topic modeling and narrative summarization tasks \citep{gilardi2023chatgpt, tornberg2023chatgpt, zhang2024benchmarking, mu2024large}. Building upon this promising evidence, this paper introduces and validates a hybrid methodological framework that leverages open-source LLMs in combination with human-in-the-loop validation. This integrated approach achieves greater interpretability and conceptual precision, effectively addressing semantic ambiguities (e.g., distinguishing between "bank" as a financial institution vs. riverbank) and contextual nuances (e.g., interpreting "liberal" as a political ideology vs. open-mindedness). Ultimately, the proposed framework facilitates robust and scalable text analysis with reduced dependence on extensive manual refinements \citep{huang2023advancing, xu2024large}.

This study aims to harness the capabilities of LLMs to propose a hybrid approach. This iterative process merges automated text analysis with human-in-the-loop validation to balance efficiency and conceptual soundness. Although LLMs demonstrate capabilities in analyzing extensive textual segments, especially by identifying nuanced narratives or sentiments within entire paragraphs, the expertise of domain specialists is crucial for refining and validating the constructs that arise. By integrating LLM-driven generation of frames, topics, or narratives with expert review, this hybrid methodology ensures both scalability and theoretical robustness. For illustration, the model might generate an initial set of frames from a large news corpus, grouping articles by themes such as economic impact or political blame. Human experts then refine or merge these themes based on domain-specific knowledge before feeding them back into the model for final classification. 

This hybrid framework integrates LLM-driven text summarization and concept generation with a human-in-the-loop validation process to improve interpretability compared to purely automated approaches in quantitative text analysis. The approach proceeds in an iterative cycle. First, the LLM generates concise summaries and tentative conceptual categories (e.g. potential frames, topics, or narratives). Researchers then refine and validate these results, ensuring alignment with domain expertise and theoretical constructs. By merging computational efficiency with expert oversight, the framework addresses one of the most significant challenges in computational social science: achieving both conceptual precision and scalability in text analysis \citep{gilardi2023chatgpt}.

Together, this paper introduces one of the first frameworks that systematically integrates LLM-based analysis with structured human input at each stage of concept identification. This synergy improves adaptability and robustness, enabling a more comprehensive identification of frames, topics, and narratives. Researchers can thus move beyond the limitations of classical models and capture the complexities present in large and evolving corpora of textual data \citep{tornberg2023chatgpt, alizadeh2023open}.

The following sections detail the theoretical foundations of the approach and review the current state of NLP techniques for concept classification. The hybrid framework is then presented, including data sources, the specifics of LLM implementation, and the staged classification process with human-in-the-loop refinement. Subsequent sections report empirical applications in multiple data sets and discuss implications for research in digital democracy, political communication, and computational social sciences \citep{gilardi2023chatgpt, lam2024concept}. The conclusion reflects on the benefits and limitations of the framework, as well as potential avenues for further refinement, including domain-specific fine-tuning and more sophisticated active learning strategies to enhance validity, interpretability, and robustness.

\section{Theoretical Framework}
\subsection{Current State of NLP Techniques for concept classification}
Applying natural language processing (NLP) techniques to measure classification constructs such as topics, frames, and narratives has seen significant advancements. Traditional methods like Latent Dirichlet Allocation (LDA), Structural Topic Models (STM), and Bertopic have been extensively used. These methods utilize statistical approaches to identify latent structures in text data, enabling the extraction of themes and topics. LDA, introduced by \cite{blei2003latent}, is a generative probabilistic model that assumes documents are mixtures of topics and topics are mixtures of words. It has been widely applied in various domains but has crucial limitations, including difficulty interpreting topics and sensitivity to the number of topics specified a priori.

Structural Topic Models (STM), proposed by \cite{roberts2014structural}, extend LDA by incorporating document-level covariates to improve the model's interpretability and flexibility. STM allows researchers to analyze how topics correlate with metadata, providing a more nuanced understanding of the text. However, STM still faces challenges in terms of scalability, computational complexity, and topic stability. Bertopic, a recent development by \cite{grootendorst2022bertopic}, leverages BERT embeddings and clustering algorithms to generate topic models. While it improves topic coherence and interpretability, Bertopic's reliance on pre-trained embeddings limits its adaptability to domain-specific language nuances. For instance, recent studies have highlighted the potential of prompt engineering techniques in LLMs to address some of these limitations, allowing for more flexible and domain-specific adaptations \citep{vatsal2024survey}.

Despite these advancements, traditional topic modeling techniques face several problems. They often produce results that require substantial manual refinement to ensure interpretability. Moreover, these models struggle with semantic ambiguity and contextual nuances, leading to topics that may not align well with the intended constructs of interest. Furthermore, these methods typically do not account for the temporal dynamics of topics, frames, or narratives, which are crucial to understanding the evolution of discourse over time. 

In addition to LDA, STM, and Bertopic, other methods, such as cluster analysis and network analysis, are increasingly used to improve the classification and interpretation of textual data. Cluster analysis groups similar data points according to their attributes and uncovers hidden patterns within large data sets \citep{jain2010data, jain1999data}. This method helps identify distinct groups or segments within text corpora, providing insight into varying themes or opinions. For example, clustering algorithms such as k-means or hierarchical clustering can be applied to text data to reveal subgroups of documents with similar thematic content \citep{aggarwal2012survey}.

Network analysis examines relationships between entities within a data set, visualizing these connections as networks or graphs. This approach is instrumental in understanding the dynamics and structures of social interactions, information flow, and influence patterns \citep{borgatti2009social}. Researchers can explore connections between different topics, actors, or concepts by applying network analysis to text data. This offers a deeper understanding of the complex interrelations in political discourse and digital communication \citep{milojevic2014network}. However, these methods also have limitations, such as the need for high-quality data and significant computational resources.  Furthermore, classifying complex constructs like frames and narratives often requires nuanced interpretation and a deep understanding of the context, which can be challenging to achieve purely through automated methods without significant human oversight.

\subsection{Capabilities of Large Language Models}
Large Language Models (LLMs) such as chatGPT and its successors and counterparts have revolutionized the field of natural language processing (NLP). These models, trained on vast amounts of text data, demonstrate remarkable abilities in understanding and generating human-like text \citep{brown2020language, zhu2023large}. One of the critical capabilities of LLMs is their proficiency in summarization and information extraction. As noted in \cite{mccoy2024embers}, this proficiency is shaped by the autoregressive nature of LLMs, which focus mainly on the prediction of the next word.  LLMs can distill lengthy documents into concise summaries, effectively capturing the core information. This is achieved through techniques such as few-shot learning, where the model is provided with examples to guide its output \citep{zhu2023large, tai2024examination}.

In addition to few-shot learning, LLMs can also perform zero-shot learning, handling entirely new tasks without any explicit training examples by relying on general textual prompts or instructions \citep{brown2020language, chew2023llm}. These approaches enable LLMs to perform complex tasks with greater accuracy and adaptability, making them powerful tools in various applications, from customer service to academic research \citep{yang2024harnessing}. Specifically, \cite{mccoy2024embers} contend that although the training of LLMs is structured around high-probability next-word prediction, this characteristic facilitates their efficient adaptation to tasks characterized by identifiable probabilistic structures. In cases where the task can be framed as a sequence of high-probability linguistic or conceptual steps, the autoregressive model excels, supporting the rationale behind leveraging LLMs for complex and nuanced tasks like frame and narrative extraction. However, the approach proposed by \cite{mccoy2024embers} also suggests that LLM success in zero-shot and few-shot tasks might be limited when faced with low-probability scenarios, which do not align with their training data distributions. In these cases, the internal heuristics of the model may not generate coherent or accurate output, leading to ``surprising failure modes''. Such failures are especially prone to occur in out-of-distribution tasks where the required reasoning or domain knowledge goes beyond the patterns captured during training.  In addition, prompt engineering methods have been explored to enhance the adaptability and specificity of LLMs for various NLP tasks, demonstrating improved performance in extracting and classifying complex constructs, such as frames and narratives \citep{vatsal2024survey}.

The summarization by LLMs takes advantage of several advanced mechanisms. First, they utilize attention mechanisms that allow the model to focus on the relevant parts of the text while ignoring the less essential details \citep{vaswani2017attention}. These mechanisms are deeply influenced by the LLM's internal structure, where the autoregressive model's reliance on next-token prediction can bias outputs towards high-probability continuations, as evidenced by \cite{mccoy2024embers}. This selective attention ensures that the generated summary encapsulates the most significant points of the source material \citep{zhang2024benchmarking}. Second, the models are pre-trained in vast and diverse corpora, providing them with extensive background knowledge and contextual understanding, which enhances their ability to generate accurate and coherent summaries \citep{dagdelen2024structured}. Third, fine-tuning specific summarization tasks further enhances their ability to create precise and contextually appropriate summaries \citep{pilault2020extractive}.

LLMs excel at zero-shot, few-shot, and chain-of-thought tasks, performing new tasks with little to no task-specific training. This is particularly beneficial in extracting relevant information from text, as demonstrated in clinical meta-analyses studies. For instance, \cite{kartchner2023zero} explored the use of ChatGPT for zero-shot information extraction from clinical trials, finding that it could accurately identify and extract pertinent data with minimal manual intervention. This ability to generalize across tasks makes LLMs handy tools in text analysis \citep{chew2023llm}.

The effectiveness of LLMs in summarization tasks can be understood through several methodological lenses. First, from a machine learning perspective, LLMs employ a transformer architecture that excels at handling sequential data and capturing long-range dependencies. Incorporating prompt engineering techniques within this architecture can further enhance the ability of the model to generate more precise and relevant summaries \citep{vatsal2024survey, vaswani2017attention}. The transformer architecture, combined with an autoregressive model's probabilistic approach, provides a framework that inherently favors high-probability coherent outputs, with this alignment between architecture and training objective underpinning LLM success in summarization and information extraction tasks \citep{mccoy2024embers}. Moreover, the self-attention mechanism allows the model to weigh the importance of different words in the input text, ensuring that the summary captures essential information while omitting irrelevant details \citep{vaswani2017attention}. Recent advances in transformer models, which are also used by LLMs in general, such as the introduction of sparse attention and memory-augmented transformers, have further enhanced their capability to manage large-scale and complex data sets efficiently \citep{zaheer2020big, lewis2020retrieval}.

Second, from a natural language processing standpoint, LLMs' ability to understand and generate human-like text is rooted in their extensive pre-training on large text corpora. This pre-training phase equips the models with a deep understanding of language syntax, semantics, and pragmatics, enabling them to generate summaries that are not only concise but also coherent and contextually appropriate \citep{dagdelen2024structured}. Using masked language modeling and next-sentence prediction during pre-training significantly improves the models' contextual prediction and coherence generation capabilities \citep{devlin2018bert, radford2019language}. This is particularly important in domains such as clinical meta-analysis, where the precision and clarity of the extracted information are paramount \citep{kartchner2023zero}.

Third, LLMs' generative nature allows them to produce flexible and adaptable summaries to various contexts. Unlike rule-based or template-based summarization methods, which can be rigid and limited in scope, LLMs can generate summaries tailored to the specific needs of the task at hand. This adaptability is crucial in dynamic fields, where the ability to quickly and accurately summarize new findings can significantly impact decision making and policy formulation \citep{pilault2020extractive}.

\subsection{Identifying Frames, Topics, Narratives and more with LLMs}
LLMs' advanced language understanding capabilities make them well suited for identifying latent constructs such as frames, topics, and narratives. As defined by \cite{entman1993framing, entman2007framing}, frames involve selecting certain aspects of reality to make them more salient in communication, thereby promoting specific interpretations and evaluations. Topics represent the thematic content of discourse, while narratives encompass structured coherent sequences of events or arguments within the text.

LLMs can efficiently detect these constructs due to their deep contextual understanding \citep{lam2024concept}. By leveraging the extensive pre-training on diverse text corpora, LLMs can discern subtle linguistic cues that indicate the presence of frames, topics, or narratives. For example, an LLM can identify framing effects by recognizing patterns in how problems are defined, causes are attributed, and solutions are proposed. Similarly, the model can extract topics by clustering semantically related content and detect narratives by following the logical flow of events or arguments.

Beyond frames, topics, and narratives, LLMs can also identify other constructs, such as sentiment, stance, and rhetorical strategies. Sentiment analysis involves determining the emotional tone of the text, while stance detection assesses the author's position on a given issue. Rhetorical strategies encompass the use of language to persuade or influence the audience, including techniques such as metaphors, analogies, and appeals to emotion. The flexibility of LLMs in capturing these diverse constructs enhances their utility in comprehensive text analysis.

These mechanisms collectively enable LLMs to perform advanced summarization tasks, capturing the essence of the text while maintaining coherence and relevance. LLMs' efficiency in summarization significantly reduces the manual effort required in processing large data sets, making them invaluable in various domains such as healthcare, finance, and legal research.

\subsection{Generative Framework for Construct Identification and Classification}
Given LLMs' capabilities, I propose a generative framework to identify latent constructs from textual data. The process begins with the extraction and summarization of relevant information from the text using LLMs. This involves generating concise summaries that capture the core arguments, perspectives, or thematic elements present in the data. These summaries serve as the basis for further analysis, facilitating the identification of frames, topics, and narratives.

The generative approach involves creating potential target classes for the constructs of interest from the generated summaries.  For example, in analyzing European Parliamentary Debates on AI, the LLM first produces concise summaries of each sentence and then generates potential frame classes such as AI Benefits, AI Risks, AI Impact on Society, or AI Impact on Work. The researchers subsequently refine these automatically proposed classes through an iterative process that combines automated model evaluations and human-in-the-loop validation. By reviewing the generated classes, experts ensure alignment with theoretical constructs and empirical relevance, merging or discarding categories as necessary. This iterative refinement improves accuracy and interpretability, yielding a distilled set of frames that better capture the nuances of the underlying data.

This framework leverages LLMs' strengths in understanding and generating text while incorporating human expertise to fine-tune the final output. Integrating LLMs with human oversight ensures that the constructs identified are theoretically sound and practically meaningful. Moreover, this approach allows for dynamic adaptation of construct identification to evolving textual data, making it particularly useful in longitudinal studies or real-time text analysis.

To conclude, although traditional NLP techniques for classifying constructs have advanced significantly, they still face notable limitations in interpretability, adaptability, and scalability. LLMs offer a promising alternative due to their superior contextual understanding and generative capabilities. Using LLMs in a generative framework, researchers can efficiently extract and identify frames, topics, narratives, and other constructs from textual data. This approach combines the strengths of machine learning and human expertise, ensuring robust and nuanced text analysis.

\section{Methods}
\begin{figure}[!htbp]
    \centering
    \includegraphics[width=0.99\textwidth]{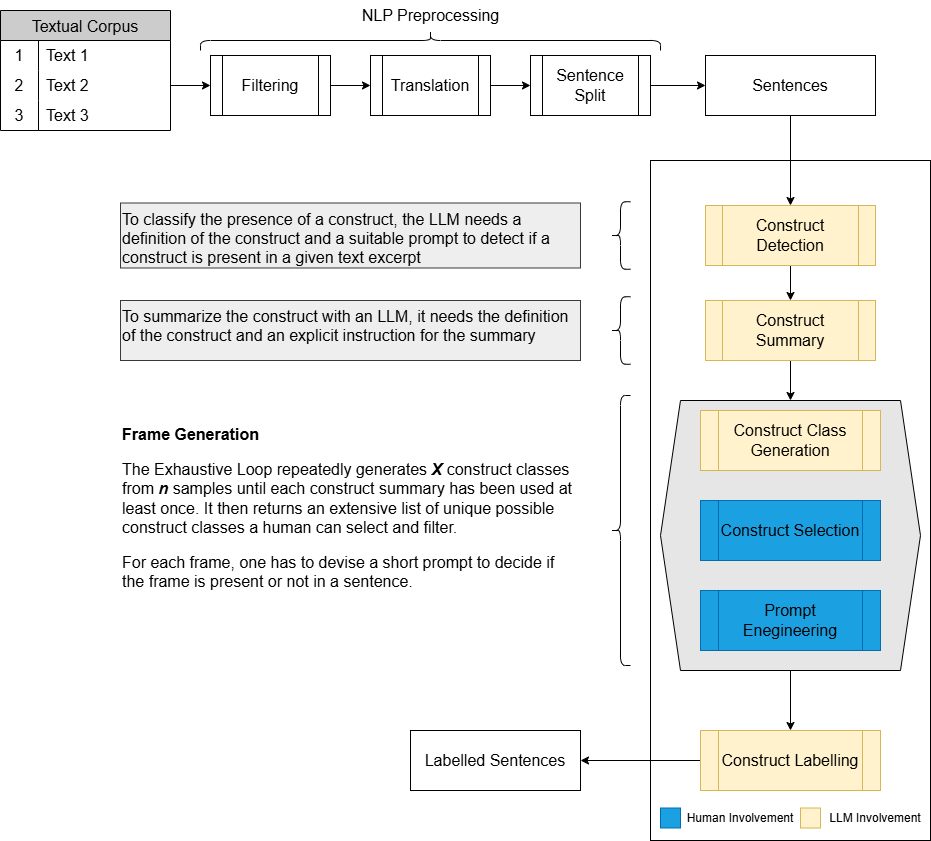}
    \caption{Construct Class Generation Framework}
    \label{method_fig_1}
\end{figure}

This paper aims to extract and classify latent constructs, including frames, narratives, and topics, from large textual corpora. By leveraging the capabilities of an open-source LLM, this approach integrates automated NLP techniques with human-in-the-loop validation, ensuring a comprehensive and reliable extraction process. The general framework is depicted in Figure \ref{method_fig_1}. 

The proposed methodological framework leverages open-source LLMs, such as the LLaMA 3 model, to extract and classify latent constructs such as frames, narratives, and topics from large textual corpora. The methodology is designed to address the limitations of traditional topic modeling methods, such as Latent Dirichlet Allocation (LDA) and Structural Topic Models (STM), which often struggle with interpretability, adaptability, and scalability \citep{blei2003latent, roberts2014structural, grootendorst2022bertopic}.

The methodological implementation of the identification of constructs in our framework proceeds through a series of clearly defined stages that demonstrate how the capabilities of the LLM are utilized in practice (see Figure~\ref{method_fig_1} for an overview). First, the texts are segmented into analytical units. Sentence-level segmentation is particularly suitable when each sentence encapsulates a distinct argument or when a high degree of granularity is required, whereas paragraph-level segmentation is more appropriate for longer texts or those containing multiple interwoven themes, balancing the need to preserve context and computational efficiency. Following segmentation, each unit is submitted to the LLM in a two-step, few-shot approach. In the initial step, the model reads the unit and produces a reasoned assessment on whether a latent construct (such as a frame or topic) is present. In the subsequent step, the unit is reprocessed, along with its initial reasoning, to confirm the decision, ensuring consistency in the detection of the construct.

Once a construct is identified, the unit is then processed with a custom summarization prompt that distills its key elements into a concise summary. It is important to note that all prompts, including those for detection, reasoning, and summarization, are carefully designed and refined iteratively by the researcher (see Appendix \ref{methods_section_appendix_prompts_frame_i_summary}, \ref{methods_section_appendix_prompts_frame_ii_summary}, and \ref{methods_section_appendix_prompts_topic_i_summary} for detailed examples of prompts). This iterative prompt engineering involves testing sample texts from the data set, evaluating whether prompts yield outputs aligned with theoretical frameworks and empirical objectives, and making necessary adjustments to optimize performance.

These summaries are aggregated iteratively through a looping procedure: In each iteration, a batch of summaries (for example, 100 at a time) is processed to generate a fixed number of candidate construct classes (for example, 10 per iteration), with each new batch partially overlapping (approximately 20 \% of the samples are repeated) with the previous one to ensure exhaustive coverage. The outcome is a comprehensive list of candidate construct classes, complete with example unit IDs and class summaries.

Subsequently, this list is refined through a human-in-the-loop step. An interactive tool displays each candidate class alongside the example units and brief construct descriptions, enabling the researcher to select only those classes that are conceptually robust and empirically pertinent. Finally, with a refined set of construct classes established, a dedicated prompt is engineered and also developed through extensive testing, to define each class for the classification phase. In this phase, every unit is evaluated on a Likert scale to determine its fit to each construct class separately. The final assignment of the construct label, or labels, is determined by integrating the quantitative fit score in an additional decision step, where the best-fitting classes are presented together for a final selection based on the prescribed instructions (see Appendix \ref{method_section_appendix_prompts_frames_i_likert}).

\subsection{Construct Identification and Summarization}
The process begins with extracting relevant information from textual data using the generative capabilities of LLMs. This involves generating concise summaries that encapsulate the core arguments, perspectives, or thematic elements present in the data. Summarization is crucial, as it distills complex texts into manageable and interpretable units, facilitating the subsequent identification of latent constructs \citep{zhu2023large}. The LLM employs a transformer architecture that utilizes self-attention mechanisms to focus on the most relevant parts of the text, enabling it to distill critical information from lengthy texts while preserving essential context and meaning \citep{vaswani2017attention}. At this stage, researchers iteratively refine the prompts used for summarization by testing them on representative portions of the data, ensuring that the outputs align with both theoretical and empirical objectives. This process, guided by human intervention, directs the LLM to prioritize salient constructs while reducing unnecessary details during subsequent analyses. This process is particularly suitable for identifying latent constructs because it reduces the cognitive load on subsequent analytical processes and improves the interpretability of complex data \citep{nenkova2012survey}. Several vital computational mechanisms underpin the effectiveness of summarization in this context. 

First, the Transformer Architecture and Attention Mechanisms introduced by \cite{vaswani2017attention} form the backbone of modern LLMs like LLaMA 3. This architecture employs self-attention mechanisms that allow the model to weigh the importance of different words and phrases within a text. By focusing on the most relevant parts of the text, the model can generate summaries that capture the core arguments and thematic elements efficiently. The self-attention mechanism is mathematically represented as: 

\begin{equation}
\text{Attention}(Q, K, V) = \text{softmax}\left(\frac{QK^T}{\sqrt{d_k}}\right)V
\end{equation}

Where \( Q \), \( K \), and \( V \) are the query, key, and value matrices, respectively, and \( d_k \) is the dimensionality of the keys. This mechanism ensures that the model considers contextual dependencies throughout the text, leading to more coherent and contextually appropriate summaries \citep{vaswani2017attention}.

Second, LLMs excel in few-shot and zero-shot learning scenarios, where they can perform new tasks with minimal task-specific training. This capability is particularly beneficial for summarization, as it allows the model to adapt to different textual domains and styles without requiring extensive retraining, as shown by various researchers \citep{kojima2022large, gilardi2023chatgpt, tornberg2023chatgpt, alizadeh2023open}. For example, in clinical meta-analyses, LLMs can accurately identify and extract pertinent information with minimal manual intervention, demonstrating their versatility and adaptability \citep{kartchner2023zero}. Moreover, prompt engineering methods can be utilized to refine LLM outputs, ensuring that the summaries are more aligned with specific research objectives and theoretical frameworks \citep{vatsal2024survey}.

Third, LLMs are pre-trained on vast and diverse corpora, providing extensive background knowledge and contextual understanding. This pre-training equips the models with a good sense of language syntax, semantics, and pragmatics, enabling them to generate summaries that are not only concise but also coherent and contextually relevant \citep{devlin2018bert}. Further fine-tuning of specific summarization tasks can further enhance their ability to produce accurate summaries tailored to the needs of particular domains or data sets, with relatively small amounts of labeled training data \citep{alizadeh2023open}.

The methodological implementation of summarization for construct identification involves several steps, each leveraging LLMs' computational strengths. It starts with text segmentation into manageable units, typically sentences or paragraphs. TThis facilitates detailed analysis and ensures that the summarization process can focus on discrete text elements. Each segment of the text undergoes a summarization using the LLM. The model generates concise summaries that encapsulate the core arguments or perspectives present in the text. The summary thus serves as the basis for identifying the latent constructs. By distilling the text, the identification of constructs is guided by pre-defined theoretical frameworks, ensuring that the extracted constructs are aligned with the research objectives and empirical relevance.

\subsection{Frame Generation and Concept Class Creation}
Once the summaries are generated, the next phase involves the iterative generation of potential target classes for the constructs of interest. For example, based on the summarized content, the model might propose frames centered on ethical debates, economic concerns, or regulatory perspectives. This generative approach leverages the LLM's ability to understand and synthesize text, producing various candidate frames that are relevant to the research objectives. The iterative process ensures comprehensive exploration of the corpus's conceptual space, meaning that all salient constructs are progressively identified. After this stage, the researchers systematically evaluate the list of suggested classes produced by the LLM, drawing on relevant theoretical frameworks and the specific scope of the study. Guided by domain expertise and empirical knowledge, they refine and finalize which classes to include, ensuring that each category reflects a conceptually sound and meaningful construct class, as explained in the following section.

Mathematically, the frame generation process can be represented as follows:

\[
F_i = g\left(\sum_{j=1}^{n} S_j\right)
\]

where \( F_i \) denotes the generated frame class, \( S_j \) represents individual summaries, and \( g \) is the generative function of the LLM.  The iterative nature of this process involves generating multiple sets of the construct of interest (e.g., frames or topics), examining them against domain-specific criteria, and refining them repeatedly, thereby exhaustively mapping the relevant conceptual space in the corpus.

\subsection{Human-in-the-Loop Validation and Refinement}
The generated concept classes then undergo a critical human-in-the-loop validation to ensure conceptual soundness and empirical relevance. The researchers review the generated classes, refining them to align with theoretical constructs and research questions. This validation step is critical to maintain the precision and interpretability of the identified constructs, addressing one of the primary limitations of fully automated methods \citep{kartchner2023zero,gilardi2023chatgpt,alizadeh2023open}. The iterative refinement process enhances model output by integrating domain expertise, thereby improving the reliability of the classification. Prompt engineering can play an expanded role in this process by providing structured prompts that not only guide the LLM toward generating outputs aligned with theoretical constructs but also incorporate domain-specific details, key definitional elements, or examples. Through such tailored instructions, the model can be nudged to focus on the most relevant aspects of the text and generate more conceptually precise frames, topics, or narratives \citep{vatsal2024survey}.

\subsection{Construct Classification}
Following the generation and validation of concept classes, the methodology employs a staged chain-of-thought approach for classification. This involves multiple stages in which the model evaluates the fit of each concept class for a given sentence or paragraph depending on the unit of analysis. Initially, the model generates a summary to encapsulate the core argument or perspective of the concept. This is followed by evaluating how well the summary aligns with the humanly validated concept classes. The model rates the fit on a scale from 1 (strongly disagree) to 7 (strongly agree) based on the coherence and relevance of the concept class to the text at hand. 

The fit evaluation can be formally expressed as:

\[
\text{Fit}(S,F) = \frac{1}{m} \sum_{k=1}^{m} \text{Rating}(S,F_k)
\]

where \(\text{Fit}(S,F)\) denotes the fit score, \(m\) is the number of frames evaluated, and \(\text{Rating}(S,F_k)\) is the rating given to the frame \(F_k\) based on its fit to the summary \(S\). The final selection of frames for each sentence involves choosing the frame(s) with the highest fit scores, ensuring a nuanced and accurate classification.

The iterative and staged nature of the method ensures greater robustness and reliability in identifying and classifying latent constructs. Furthermore, the inclusion of steps where the LLM is allowed to freely give reasoning should overall increase the performance of the frameworks, as shown by \citet{tam2024let}. By combining automated summarization and frame generation with human validation, the approach mitigates misclassification risks and enhances the interpretability of the results. Integrating human expertise at multiple stages of the process ensures that the final result is both theoretically sound and empirically relevant, addressing the challenges faced by traditional NLP methods \citep{chew2023llm}. It is also critical to ensure the interpretability and transparency of the summarization process, particularly in domains where the accuracy and clarity of extracted information are paramount \citep{pilault2020extractive}.

\section{Data}
\subsection{Frame Analysis on European Parliamentary Speeches} \label{sec_method_data_1}
To illustrate and assess the effectiveness of the proposed methodology, this study uses a comprehensive data set of EU Parliamentary debates that mention ``AI'' or ``Artificial intelligence'' in the title or within the content of the debate. The data set comprises 133 debates that took place between 2000 and 2023. The debates were collected using a web scraper that extracted relevant data from official parliamentary records. The EU Parliamentary data set was selected because it provides detailed and highly structured debates, making it well suited for testing frame and narrative identification. The data set allows for fine-grained analysis at the sentence level, capturing subtle differences in framing within the same debate. However, the data set focuses on a specific institutional and political context, which can limit the generalizability of the findings to other types of political discourse, such as informal or less structured discussions on social networks. The analysis is performed at the sentence level. This level of granularity ensures that each statement can be classified independently, allowing the detection of subtle differences in framing that might otherwise be overlooked when analyzing entire paragraphs or full speeches.

For linguistic processing, I translated speeches in Polish, Czech, Greek, Dutch, Romanian, Hungarian, Danish, Swedish, Slovakian, Finnish, Croatian, Lithuanian, Bulgarian, Estonian, and Slovenian into English using the Google Translation API. Google Translate is a statistical translation tool that calculates the probabilities of various correct phrase translations rather than focusing on word-for-word translation \citep{johnson2017google}. Speeches originally in English, Spanish, French, Italian, German, and Portuguese were analyzed in their native languages. Languages with less than 0.5\% representation in the data set and those presenting translation challenges with the Google Translation API --- Bosnian, Irish (Gaelic), Latvian, Maltese --- were excluded from the analysis.

The final data set contains 133 debates and 5,019 speeches with 19,538 sentences. This comprehensive data set enables an in-depth analysis of parliamentary debates on artificial intelligence in multiple languages and years within the European Parliament.



\subsection{Frame Analysis on Newspaper Articles on Encryption} \label{sec_method_data_2}
To demonstrate the generalizability of the proposed methodological framework, I also applied it to a data set of US newspaper articles that are relevant to the discussion concerning encryption. This data set, drawn from multiple major outlets over 23 years (2000–2023), comprises 10,954 articles that collectively capture the discourse on encryption in US news media. The encryption data set was chosen for its rich and diverse coverage of a contested policy issue over a long period of time, enabling exploration of evolving narratives and frames. However, as a media-based data set, it can be biased by editorial perspectives and news selection processes, potentially limiting its representativeness of public opinion. Moreover, its focus on US media limits cross-national comparisons, which could be an area for future work. 


\subsection{Topic Modeling with 20 Newsgroup data set} \label{sec_method_data_3}
For the topic modeling aspect of this study, the widely recognized 20 Newsgroup data set is used as a benchmark to evaluate the quality of the topic modeling algorithms. This data set contains 18,846 news articles, evenly distributed across 20 different categories, providing balanced coverage ideal for validating topic models at scale.

While inspecting the corpus, I discovered 115 articles exceeding 4,096 tokens, partly due to extensive code, HTML fragments, XML headers, or other non-standard text elements. Since LLaMA 3 has a maximum context window of 4,096 tokens, summarization could not be guaranteed for texts that surpass this limit. Moreover, 53 of these long articles originate from the category ``comp'' (computer), featuring large sections of raw hex code or HTML artifacts that contribute little to meaningful textual content. As a result, these 115 articles were excluded from the analysis. 

Unlike the frame analysis data set, the 20 Newsgroup data set allows for validation using the entire data set due to its comprehensive labeling and established use as a benchmark, for topic modeling. This enables a thorough evaluation of the topic modeling methods without the constraints of sample-based validation, thus enhancing the robustness and credibility of my findings. 

\section{Results}
\subsection{Validation of Classification Approach}
To establish robust benchmarks for evaluating the performance of LLM in the three data sets, a unified human-in-the-loop validation procedure was performed. For each data set, two trained research assistants independently coded a stratified random sample: 1,250 sentences from the European Parliamentary Debates data set, 600 paragraphs from the US News Articles on Encryption data set, and 1,000 articles from the 20 Newsgroups data set. After resolving disagreements through filtering for coder agreement, usable samples for subsequent validation were 996 sentences (Parliamentary Debates), 335 paragraphs (News Articles on Encryption), and 675 articles (20 Newsgroups).

\begin{table}[h]
    \centering
    \small
    \begin{tabular}{llccc}
        \toprule
        \textit{Data Set} & \textit{Classification Task}  & \textit{Accuracy} & \textit{Krippendorff's Alpha}\\
        \midrule
        EU Parliamentary Debates & Frame Presence & 0.89 & 0.73 \\
        EU Parliamentary Debates & Frame Classification & 0.83 & 0.60 \\
        US Articles on Encryption & Frame Presence & 0.90 & 0.65 \\
        US Articles on Encryption & Frame Classification & 0.56 & 0.51 \\
        20 Newsgroups & Topic Classification & 0.68 & 0.66 \\
        \bottomrule
    \end{tabular}
    \caption{Consolidated intercoder reliability metrics for human coders in all data sets. The reported metrics include accuracy and Krippendorff’s Alpha.}
    \label{method_table_human_coder_metrics_combined}
\end{table}

Intercoder reliability metrics calculated with Krippendorff's Alpha varied across data sets because of differences in textual granularity and complexity. Parliamentary debates exhibited strong reliability for the presence of frames (0.73) and moderate reliability for the classification of specific frames (0.60). Reliability in the encryption article data set was moderate both for the presence of the frame (0.65) and for the classification of specific frames (0.51). Similarly, topic classification within the 20 Newsgroups data set showed moderate intercoder agreement with an accuracy of 0.68 and Krippendorff's Alpha of 0.66, reflecting inherent challenges in differentiating closely related topics. These consolidated human coder reliability metrics across data sets are summarized in Table~\ref{method_table_human_coder_metrics_combined}.

The results of the human coders established clear performance benchmarks, enabling a systematic evaluation of the LLM-based classification method described in the following subsections.

\subsection{Frame Classification on European Parliamentary Debates}
The proposed methodology was applied to the European Parliamentary Debates data set described in Section \ref{sec_method_data_1}. Initially, LLM identified 83 potential frames related to AI, such as \textit{AI Benefits}, \textit{AI Risks}, and \textit{AI Ethics}. Recognizing overlaps and less relevant frames, a refinement process was conducted through human review, focusing on theoretical relevance and clarity. This led to a distilled list of 11 distinct frames: \textit{AI Benefits}, \textit{AI Risks}, \textit{AI Ethics}, \textit{AI Regulation}, \textit{AI Impact}, \textit{AI Innovation}, \textit{AI Development}, \textit{AI Potential}, \textit{AI Limitations}, \textit{AI Concerns}, and \textit{No Frame}.

The LLM was then applied to classify the same sentences. The performance metrics of the model are shown in Table~\ref{method_table_llm_metrics_ai}. The LLM achieved an accuracy of 0.84 for detecting the presence of any frame, which closely matches human performance. For classifying specific frames, the model reached an accuracy of 0.79 and an F1 score of 0.73. These results indicate that the LLM effectively identified and classified the frames within the debates.

\begin{table}[h]
    \centering
    \small
    \begin{tabular}{lcccc}
        \toprule
         & Accuracy & F1 Score & Precision & Recall \\
        \midrule
        Frame Presence & 0.84 & 0.88 & 1.00 & 0.78 \\
        Frame Classification  & 0.79 & 0.73 & 0.40 & 0.74 \\
        \bottomrule
    \end{tabular}
    \caption{LLM performance metrics in classifying frames within the debates. The table presents accuracy, F1 score, precision, and recall for detecting the presence of any frame and classifying specific frame classes, demonstrating the model's effectiveness compared to human coders.}
    \label{method_table_llm_metrics_ai}
\end{table}

The integration of human-in-the-loop experience in the refinement of the frame classes proved crucial. The LLM's performance can be improved significantly after the frames are refined. This reduces ambiguity and overlap. This underscores the importance of combining automated LLM capabilities with human-in-the-loop validation to enhance both the accuracy and interpretability of classification results.

These findings support the argument that integrating LLM outputs with human expertise enhances the quality and accuracy of frame extraction and classification. The human-in-the-loop process improved the clarity of frames, boosting the LLM's classification performance. This approach addresses challenges in traditional NLP methods by combining scalability with conceptual soundness.

In summary, this first example demonstrates that LLMs, when guided by human validation, can effectively extract and classify frames from large textual data sets such as parliamentary debates. This offers a valuable tool for researchers analyzing complex policy discussions.

\subsection{Frame Classification on News Articles covering Encryption}
The second data set focuses on how encryption is framed within the US, providing a second opportunity to test the effectiveness of the proposed method in a different setting. The initial output of the model includes an extensive list of frames, some of which overlap or are not directly relevant to the focus on encryption as a single issue topic. This underscores the need for careful human oversight to ensure that the selected frames are conceptually distinct and relevant to the debate on encryption. Through iterative refinement, one can distill the list down to a set of mostly mutually exclusive frames, avoiding redundancy while capturing the diverse ways encryption is discussed in the media. The main advantage of this frame generation process is that it provides a researcher with an exhaustive list of possible frames measured in the data. This allows for a more manageable selection of a set of frame classes without missing important frames that could be left out. 

This refinement process leads to the selection of frame classes that are both theoretically grounded and empirically relevant, enabling a nuanced analysis of how encryption is framed in the news landscape of the USA. The final set of frame classes provides a coherent structure for examining encryption's competing narratives and priorities, including security, privacy, ethics, public safety, and moral evaluations. The generative process consistently suggested these frames across multiple sampling iterations, confirming their empirical relevance and theoretical grounding.

Building on the model's suggestions and refining them further, I identified eleven key frame classes to analyze how encryption is portrayed in the news media. Governmental Control, Corporate Power, Accountability Issues, Privacy vs. Security, Privacy, Security, Surveillance Concerns, National Security, Encryption Risks, Encryption Threats, and Other Frames. These categories capture the range of perspectives and narratives surrounding encryption, allowing a systematic examination of media frames that reflects the complexities and nuances of public discourse on this topic. 

Evaluated against human coder benchmarks (Table~\ref{method_table_llm_metrics_encr}), the LLM demonstrated strong accuracy (0.92) in detecting the presence of frames and moderate accuracy (0.66) with an F1 score of 0.68 for classifying specific frame classes. Higher precision (0.61) relative to recall (0.52) highlights the inherent challenges posed by complex journalistic discourse.

\begin{table}[h]
    \centering
    \begin{tabular}{lcccc}
        \toprule
         & Accuracy & F1 Score & Precision & Recall \\
        \midrule
        Frame Class Present & 0.92 & 0.70 & 0.75 & 0.66\\
        Frame Classes & 0.66 & 0.68 & 0.61 & 0.52\\
        \bottomrule
    \end{tabular}
    \caption{LLM performance metrics in classifying frames within the paragraphs. The table presents accuracy, F1 score, precision, and recall for detecting the presence of any frame and classifying specific frame classes, demonstrating the model's effectiveness compared to human coders.}
    \label{method_table_llm_metrics_encr}
\end{table}

The results reinforce the importance of combined human-LLM refinement in the effective management of nuanced content.

\subsection{Topic Classification}
To evaluate the reliability of the method beyond frame analysis, I applied it to the 20 Newsgroups data set. This data set enables me to assess the method's capability to classify documents into distinct topics. Similarly to the other data sets, the primary output of interest is the generated list of concepts, which in this context corresponds to topics. The generation process identified a total of 20 unique topics. As with the other data sets, many of these topics are closely related. This allows for a thorough evaluation of the method's ability to detect all the topics and the frequency with which each topic is identified in the samples.

Of the 20 annotated topics, the generative process successfully identified 11 topics, the most frequently observed being Christianity, Cryptography, Hockey, Space Exploration, Baseball, Motorcycles, Politics, Medicine, Religion, Automotive, and Technology. Several other topics, such as Social Issues, International Relations, Military and Defense, Law, Gaming, Philosophy, Education, Business, Science, Health, History, and general Sports, were not directly identified. However, these topics could be partially inferred through closely related categories identified in the extended list of generated topics.

The additional topics generated, such as ``Law,'' ``Gaming,'' ``Philosophy,'' ``Education,'' ``Business,'' ``Science,'' and ``Health,'' were categorized under broader headings like ``Politics and Government,'' ``Technology and Computing,'' and ``Religion and Philosophy.'' This highlights the method's ability to capture nuanced variations within broader thematic areas. For example, ``Military and Defense'' could be inferred under ``Security,'' and ``International Relations'' might be partially represented by topics such as ``Conflict and War'' or ``International Relations'' in the generated list.

Overall, the method demonstrates a solid ability to identify most topics, although it encounters some difficulty with closely related topics that may function as subtopics of others. However, the topic-generation process is effective and even proposes additional sensible topics that extend the initial set, enhancing the overall comprehensiveness of the topic-identification process. Ultimately, I settled on 25 topics, including a miscellaneous one to catch all texts that my selection could not classify. Below are the topics and their corresponding explanations, along with the original topics from the 20 Newsgroups data set they are related to:

\begin{itemize}[itemsep=-1ex]
    \item SOCIAL ISSUES \emph{(Related to: talk.politics.misc, talk.politics.guns)}
    \item INTERNATIONAL RELATIONS \emph{(Related to: talk.politics.mideast)}
    \item MILITARY AND DEFENSE \emph{(No direct match, partly from talk.politics.guns and talk.politics.mideast)}
    \item CHRISTIANITY \emph{(Direct match: soc.religion.christian)}
    \item LAW \emph{(No direct match, partly from talk.politics.misc and talk.politics.guns)}
    \item GAMING \emph{(No direct match)}
    \item CRYPTOGRAPHY \emph{(Direct match: sci.crypt)}
    \item PHILOSOPHY \emph{(No direct match, partly from alt.atheism and talk.religion.misc)}
    \item HOCKEY \emph{(Direct match: rec.sport.hockey)}
    \item SPACE EXPLORATION \emph{(Direct match: sci.space)}
    \item BASEBALL \emph{(Direct match: rec.sport.baseball)}
    \item MOTORCYCLES \emph{(Direct match: rec.motorcycles)}
    \item EDUCATION \emph{(No direct match, partly from sci.med and sci.electronics)}
    \item POLITICS \emph{(Direct match: talk.politics.misc)}
    \item MEDICINE \emph{(Direct match: sci.med)}
    \item BUSINESS \emph{(No direct match, partly from misc.forsale and some discussions in rec.autos and rec.motorcycles)}
    \item SCIENCE \emph{(No direct match, partly from sci.electronics, sci.med, and sci.space)}
    \item HEALTH \emph{(No direct match, partly from sci.med and misc.forsale)}
    \item HISTORY \emph{(No direct match)}
    \item RELIGION \emph{(Direct match: talk.religion.misc, alt.atheism)}
    \item AUTOMOTIVE \emph{(Direct match: rec.autos)}
    \item SPORTS \emph{(No direct match, but partly from rec.sport.xy classes)}
    \item TECHNOLOGY \emph{(Direct match: all comp.xy classes)}
    \item MISCELLANEOUS \emph{(No direct match)}
\end{itemize}

To assess the classification performance, two human coders independently classified a random sample of 1,000 articles, resulting in 675 usable articles after resolving coding disagreements. They achieved an accuracy of 0.68 and a Krippendorff's Alpha of 0.66, as summarized in Table~\ref{method_table_human_coder_metrics_combined}.

Against the human benchmark, the LLM achieved a comparable accuracy of 0.65 and an F1 score of 0.68, despite notably lower precision (0.33), suggesting conservative classifications due to overlapping semantic content. These metrics (see Table~\ref{method_table_llm_metrics_topics}) indicate the method’s overall effectiveness in handling nuanced topics at scale.

\begin{table}[h]
    \centering
    \begin{tabular}{lcccc}
        \toprule
         & Accuracy & F1 Score & Precision & Recall \\
        \midrule
        Topics  & 0.65 & 0.68 & 0.33 & 0.60 \\
        \bottomrule
    \end{tabular}
    \caption{Performance metrics (precision, recall, and F1 score) of the LLM in classifying articles into 25 LLM-derived topics. These metrics illustrate the model's performance relative to human coders in the topic classification task.}
    \label{method_table_llm_metrics_topics}
\end{table}

These results indicate that the LLM effectively classifies topics even when using a set of topics generated by the method itself. The moderate agreement levels suggest that the task is inherently challenging due to the nuanced and overlapping nature of some topics. Nonetheless, the LLM's performance is comparable to that of human coders, highlighting its potential as a valuable tool for topic classification in large textual data sets.

In the classification process, the LLM employs a staged chain-of-thought approach as outlined in the methodology. Initially, it generates a concise summary of each article or paragraph to distill the essential content related to potential topics. This summarization captures the core arguments and thematic elements of the text. The LLM then evaluates the fit of each possible topic class to the summarized content by rating the coherence and relevance of each topic class on a scale from 1 (strongly disagree) to 7 (strongly agree). The model selects the topic class with the highest fit score as the best-fitting topic for that text segment. This systematic approach enhances classification accuracy by focusing on the most relevant aspects of the text and leveraging the LLM's contextual understanding.

Applying the methodology to the 20 Newsgroups data set demonstrates its adaptability and effectiveness in topic classification tasks. By integrating LLM-generated topics with human validation, we ensure both the comprehensiveness and relevance of the topic categories. Despite the inherent challenges of nuanced and overlapping topics, the LLM performs comparably to human coders. This underscores the benefit of combining automated processing with human expertise in text analysis, offering a robust framework for topic classification in large and complex data sets.

\subsection{Comparison of Classification Results Across Data Sets}
This section compares the performance of LLM in three classification tasks, as summarized in Table \ref{method_table_llm_classification_comparison}. Each data set features a different structure and content type, which influences the accuracy, precision, recall, and F1 scores of the LLM.

The European Parliamentary Debates data set consists of speeches analyzed at the sentence level for frame classification. As shown in Table \ref{method_table_llm_classification_comparison}, LLM achieved its highest accuracy (0.79) and a solid F1 score (0.73). These results suggest that the formal and policy-focused nature of parliamentary speeches helps the model identify frames more consistently. However, precision (0.40) was significantly lower than recall (0.74), indicating that at sentence-level granularity, the model sometimes overgeneralized and struggled to distinguish closely related frames clearly.

\begin{table}[h]
    \centering
    \begin{tabular}{lcccc}
        \toprule
        \textbf{Data Set} & \textbf{Accuracy} & \textbf{F1 Score} & \textbf{Precision} & \textbf{Recall} \\
        \midrule
        \textit{European Parliamentary Debates} & 0.79 & 0.73 & 0.40 & 0.74 \\
        \textit{US News Articles on Encryption} & 0.66 & 0.68 & 0.61 & 0.52 \\
        \textit{20 Newsgroups (Topic Classification)} & 0.65 & 0.68 & 0.33 & 0.60 \\
        \bottomrule
    \end{tabular}
    \caption{Performance metrics of the LLM across different classification tasks. The table presents accuracy, F1 score, precision, and recall for frame classification in \textit{European Parliamentary Debates} and \textit{US News Articles on Encryption}, as well as topic classification in the \textit{20 Newsgroups} data set.}
    \label{method_table_llm_classification_comparison}
\end{table}

In contrast, classifying frames in the US News Articles on Encryption data set proved to be more challenging. This task involved a paragraph-level classification of a topic that tends to be more fragmented and varied in tone. The model achieved moderate accuracy (0.66) and an F1 score of 0.68, reflecting decent but not outstanding performance. Unlike parliamentary debates, news coverage can merge different viewpoints in a single paragraph, which contributed to the slightly higher precision of the model (0.61) but lower recall (0.52). Essentially, the model became more selective in assigning frames, correctly classifying fewer cases and thus missing some relevant instances. 

However, classifying at the paragraph level can capture a broader context that may be overlooked in a sentence-level approach, potentially revealing the most prominent or overarching constructs that span multiple sentences. This trade-off means that while precision and recall may vary, coarser segmentation can also provide richer insights into how a single unit of text weaves together multiple arguments or frames.

Finally, the 20 Newsgroups data set required topic (rather than frame) classification across entire texts. According to Table \ref{method_table_llm_classification_comparison}, LLM scored an accuracy of 0.65 and an F1 score of 0.68 --- comparable to the encryption data set. However, the precision dropped substantially to 0.33, highlighting the difficulty in distinguishing semantically similar topics. Because each text often touched on multiple themes, the model sometimes struggled to reliably assign the correct topic category.

These findings generally emphasize that LLM-based classification is strongly influenced by text structure and granularity. Sentence-level tasks in structured political speeches offer reliable results, whereas paragraph-level journalistic content and full-text topical discussions introduce additional ambiguity. However, larger segments can help researchers identify high-level constructs that emerge across multiple sentences, at the cost of finer-grained precision. Further refinement of frame and topic definitions, along with domain-specific fine-tuning and human validation, can help address challenges like conceptual overlaps and multi-themed texts. Although LLM demonstrates considerable promise in automated classification, Table \ref{method_table_llm_classification_comparison} highlights how performance varies when faced with different linguistic and contextual complexities.

\section{Conclusion}
This study introduces a novel hybrid framework for latent construct extraction and classification, leveraging the capabilities of open-source Large Language Models (LLMs) such as LLaMA 3. This framework combines generative capabilities with human-in-the-loop validation to overcome key limitations of existing techniques like Latent Dirichlet Allocation (LDA) and Bertopic. Unlike these traditional methods, it is better suited for extracting and classifying latent constructs, such as frames, topics, and narratives, while offering greater conceptual alignment and adaptability to specific research questions. This flexibility allows researchers to measure precise constructs tailored to a given study's theoretical and empirical needs, advancing the analytical rigor of computational social science.

The framework's application to data sets spanning European Parliamentary debates, US news articles on encryption, and a benchmark topic modeling corpus demonstrates its robustness and versatility. The framework improves scalability and accuracy by generating explicit, consistent constructs while performing well in frame- and topic-classification tasks. While other methods often struggle with interpretability, ambiguous word meanings, and the evolving nature of textual data, the LLM-based approach addresses these challenges by using contextual embeddings and iterative refinement processes. This positions the framework as a valuable tool for addressing complex political communication, media studies, and public policy discourse.

However, it is important to acknowledge the trade-offs inherent in this approach. Despite its scalability and adaptability, the method still does not fully match the precision of expert human coders. Human validation remains crucial to ensure conceptual validity and resolve uncertainties in construct identification, underscoring the current limitations of automated processes. Although some researchers may rely on dedicated research assistants for this validation step, crowd-sourcing platforms (e.g., Amazon Mechanical Turk, Prolific) also offer feasible alternatives for recruiting human annotators at scale \citep{gilardi2023chatgpt}. This flexibility in how human validation is conducted broadens the feasibility of the method. Overall, the framework represents a significant step toward balancing scalability with accuracy, making it particularly useful for large-scale studies where exclusively manual coding would be prohibitively resource-intensive.

Future research should enhance the framework's performance by integrating fine-tuning techniques with small, domain-specific training sets. Fine-tuning open-source LLMs to align more closely with the constructs and contexts of interest can significantly improve classification outcomes while retaining the adaptability of the generative approach. This refinement could mitigate performance gaps between machine-driven and human coding, enabling more reliable and nuanced analysis across diverse data sets.

Additionally, the methodology's reliance on LLMs underscores the importance of ethical and epistemological considerations. Ensuring transparency in decision making, addressing potential biases in pretrained models, and maintaining interpretability are crucial for upholding scientific rigor. Future advancements should explore semiautomated validation mechanisms and active learning strategies to further streamline human-in-the-loop processes without compromising conceptual integrity.

In summary, this study demonstrates that combining the scalability and adaptability of LLMs with human expertise represents a transformative advancement over traditional methods such as LDA and Bertopic modeling. By enabling tailored and precise construct identification and classification, the framework offers researchers a powerful tool for analyzing large and complex textual data sets. Although not yet a complete replacement for expert human coders, its scalability and conceptual alignment make it an invaluable addition to the methodological toolkit of computational social scientists. With continued refinement and integration of fine-tuning techniques, this hybrid approach holds great promise for advancing our understanding of the nuanced layers of meaning embedded in textual discourse, ultimately broadening the scope and depth of social science research.

\paragraph{Acknowledgment}
I thank Daria Stetsenko, Dina Della Casa, and Benjamin Streiff for their excellent research assistance. I am grateful to Emma Hoes and Fabrizio Gilardi for their valuable feedback on earlier versions of this paper.

\paragraph{Funding Statement}
This project received funding from the European Research Council (ERC) under the European Union's Horizon 2020 research and innovation program (grant agreement nr. 883121).


\bibliography{references}  

\newpage
\appendix
\renewcommand{\thesection}{S\arabic{section}}
\setcounter{figure}{0} \renewcommand{\thefigure}{S\arabic{figure}}
\setcounter{table}{0} \renewcommand{\thetable}{S\arabic{table}}

\section{Human-in-the-Loop Protocol}
Human input remains central to this framework for both conceptual and practical purposes. On a theoretical level, it helps confirm that the concepts identified by the LLM match the foundational principles and empirical objectives of the project. Although automated systems are powerful, they can overlook subtle distinctions, merge overlapping categories, or miss context-specific insights. By relying on expert judgment, researchers maintain the clarity and thematic consistency that fully automated methods often struggle to maintain. From a practical point of view, this involvement occurs through a step-by-step process in which the researcher iteratively refines the questions, reviews the outputs, and adjusts both the definition of these constructs and the classification guidelines.

During the first phase, the researcher focuses on prompt engineering for the detection of constructions. This entails drafting and revising prompts to direct the LLM to surface relevant frames, topics, or narratives. Textual samples from the corpus are analyzed one at a time, and the researcher systematically adjusts instructions or reformulates queries based on whether the LLM’s responses match the intended conceptual boundaries. If the model misses pertinent constructs or conflates distinct concepts, the researcher refines the language of the prompts, taking into account both the theoretical framework and the observed behavior of the model. In practice, this could mean repeatedly testing and updating prompts until the output begins to consistently capture the latent constructs essential to the study.

Once the model reliably identifies the constructs, the researcher introduces an additional layer of prompt engineering aimed at producing coherent and concise summaries of each detected concept. Here, human oversight is crucial for checking whether these summaries highlight key arguments, omit extraneous detail, and preserve contextual accuracy. The researcher may compare multiple versions of a summary generated with different prompts, decide which one best captures the theoretical core of the construct, and then finalize the template prompt. This iterative approach helps ensure that the results are both comprehensible and aligned with the analytical objectives of the study.

With the constructs clearly identified and summarized, the final step involves creating a stable set of classes for large-scale classification. At this stage, the LLM first produces a comprehensive list of candidate classes by systematically scanning the entire corpus of texts which exhibit the construct of interest, whether a frame, narrative, or topic. This exhaustive survey yields a table that shows each proposed category, illustrated with example IDs, and an indication of how frequently that category occurs. The researcher then reviews these conceptual suggestions, checking that each class aligns with the theoretical framework of the project and does not overlap excessively with other classes. If certain categories are ambiguous, too narrowly defined, or conceptually redundant, the researcher merges or discards them. After selecting the most coherent set of categories, the researcher refines the prompts once again so that the LLM can consistently rate if a construct class fits to the text instances that exhibit the construct in question. Before finalizing the pipeline, the researcher tests this classification approach on a smaller sample of texts, examining both the assigned labels and any systematic errors that emerge. If necessary, further adjustments are made to the prompt or the class structure. These refinements are based on the in-depth understanding of the researcher's corpus, ensuring that the final classification scheme accurately represents the complexity and thematic diversity of the analyzed texts.

Following these iterative refinements, the last safeguard is a pass of verification by RAs. Their review takes place only after the classification pipeline is fully developed, introducing a new set of eyes and minimizing any bias introduced by the iterative participation of the researcher.  RAs assess a sample of classifications to confirm that it meets established conceptual standards and genuinely represents the constructs defined earlier in the process. By combining in-depth theory-driven prompt development with an impartial final validation, the framework balances specialized domain expertise against the need for independent quality assurance, ultimately delivering more reliable and interpretable results at scale.

\section{Prompts}
\subsection{Prompts Frame Analysis I} \label{method_appendix_prompts_f1}
This section outlines the framework used for frame generation and classification in the context of parliamentary speeches on artificial intelligence. The process uses an LLM in a sophisticated chain of functions filtering sentences with frames, summarizing the frames, generating frame classes out of these summaries, and finally labeling the frames within the sentences after evaluating and selecting suggested frame classes by the author. 

\subsubsection{Frame Detection}
The first stage of the framework is identifying the frames in sentences. This involves a chain-of-thought interaction with the LLM, resulting in a binary classification that labels the presence or absence of a frame.

\begin{tcolorbox}[colframe=black!75!white, colback=white, boxrule=0.5mm, width=\textwidth, sharp corners, left=1mm, right=1mm, top=1mm, bottom=1mm]
\begin{onehalfspacing}
\textbf{Interaction One:}

\textbf{System:} 

Determine whether the following sentence (not the title) from a parliamentary speech on artificial intelligence exhibits a frame (framing effect).

According to Robert Entman's definition where framing suggests that frames select some aspects of a perceived reality and make them more salient in a communicating text, 
in such a way as to promote a particular problem definition, causal interpretation, moral evaluation, and/or treatment recommendation for the item described.

Assess if specific language choices, focus points, or implied assumptions in the sentence promote a distinct perspective or argument, indicative of a frame.
\vspace{1em}

\textbf{User:} 

Debate Title: [Title ] 

Text: [SENTENCE]

Is a frame present regarding artificial intelligence (yes, no)? Provide some thoughts.
\vspace{1em}

\textbf{LLM Answer:} 

[THOUGHTS]
\vspace{1em}
\end{onehalfspacing}
\end{tcolorbox}

\begin{tcolorbox}[colframe=black!75!white, colback=white, boxrule=0.5mm, width=\textwidth, sharp corners, left=1mm, right=1mm, top=1mm, bottom=1mm]
\begin{onehalfspacing}
\textbf{Interaction Two:}

\textbf{System:} 

Determine whether the following sentence (not the title) from a parliamentary speech on artificial intelligence exhibits a frame (framing effect).

According to Robert Entman's definition where framing suggests that frames select some aspects of a perceived reality and make them more salient in a communicating text, 
in such a way as to promote a particular problem definition, causal interpretation, moral evaluation, and/or treatment recommendation for the item described.

Assess if specific language choices, focus points, or implied assumptions in the sentence promote a distinct perspective or argument, indicative of a frame.
\vspace{1em}

\textbf{User:} 

Debate Title: [Title] 

Text: [SENTENCE] 

Is a frame present regarding artificial intelligence (yes, no)? Provide some thoughts.
\vspace{1em}

\textbf{Assistant:} 

[THOUGHTS]
\vspace{1em}

\textbf{User:} 

Now answer with just the correct category (yes, no). Please answer with just Yes or No.
\vspace{1em}

\textbf{LLM Answer:} 

[LABEL] (Yes, or No)
\vspace{1em}
\end{onehalfspacing}
\end{tcolorbox}

\subsubsection{Frame Summarizing} \label{methods_section_appendix_prompts_frame_i_summary}
All sentences with a frame are then summarized with the LLM via Zero-Shot. This involves a system prompt and a user prompt, which guide the large language model to generate concise summaries of the frames found in a sentence. The system prompt includes a definition of a frame and an explanation of the task, instructing the LLM to summarize the sentence with a frame in no more than ten words.

\begin{tcolorbox}[colframe=black!75!white, colback=white, boxrule=0.5mm, width=\textwidth, sharp corners, left=1mm, right=1mm, top=1mm, bottom=1mm]
\begin{onehalfspacing}
\textbf{System:} 

Analyze the sentence provided and identify any frame present.

According to Robert Entman's definition where, framing suggests that frames select some aspects of a perceived reality and make them more salient in a communicating text, 
in such a way as to promote a particular problem definition, causal interpretation, moral evaluation, and/or treatment recommendation for the item described.

Assess if specific language choices, focus points, or implied assumptions in the sentence promote a distinct perspective or argument, indicative of a frame.

Summarize the identified frame in no more than 10 words.
\vspace{1em}

\textbf{User:} 

Please write a summary of the argumentative, indicative, and or conceptual frame about AI present in the following sentence in 3 to 10 words: [SENTENCE]

Make sure to ensure the summary captures the core argument or perspective highlighted in the sentence.
\vspace{1em}

\textbf{LLM Answer:} 

[FRAME SUMMARY]
\vspace{1em}
\end{onehalfspacing}
\end{tcolorbox}

\subsubsection{Frame Generation}
To generate all possible frame classes, the frame generation process uses a series of zero-shot interactions together with the previously obtained frame summaries to generate frame class suggestions. This involves exhaustively drawing 50 summaries simultaneously and generating a maximum of 9 frame classes per call. This approach ensures a comprehensive list of possible frames, as each frame summary is seen at least once, with 20 \% of the 50 retained for the next call, while 40 new ones are added. Hence, the process begins by initializing the necessary variables and setting configurations for the LLM. Then, the loop iterates over the data frame containing frame summaries, selecting and sampling summaries to generate frame class suggestions. Thus, the system prompt includes the task and a definition of the framing concept, instructing the LLM to categorize the frame summaries into no more than 9 classes. The prompt provides the specific instructions and the summaries to the LLM, ensuring the response includes the frame class name and count of summaries in a machine-readable JSON format. 

\begin{tcolorbox}[colframe=black!75!white, colback=white, boxrule=0.5mm, width=\textwidth, sharp corners, left=1mm, right=1mm, top=1mm, bottom=1mm]
\begin{onehalfspacing}
\textbf{System:} 

Categorize the following set of frame summaries from a set of sentences with a frame present.
According to Robert Entman's definition, framing involves selecting and emphasizing aspects of a situation or issue to promote specific problem definitions, causal interpretations, moral evaluations, and treatment recommendations.

This shapes the social reality and guides the audience's understanding.

Please come up with a set of categories but no more than 1 to 9 for these examples.
\vspace{1em}

\textbf{User:} 

I have these frame summaries:

[FRAME SUMMARIES]

Please come up with a maximum 9 frame classes and write a 2-4 word name for the frame category, the number of times this frame occurs in this sample, and an associated chatGPT Prompt that can serve as an instruction to determine if a new sentence contains this frame (framing effect) or not.

Please respond ONLY with a valid JSON in the following format (No yapping!):

\begin{lstlisting}
{
  "frame-categories": [
    {
        "frame": "<FRAME_NAME_1>", 
        "prompt": "<FRAME_NAME_PROMPT_1>", 
        "Count": "<NUMBER_OF_SUMMARIES_1>", 
        "Example IDs": "<URN_ID_1_1, URN_ID_1_2, URN_ID_1_3>"
    },
    {
        "frame": "<FRAME_NAME_2>", 
        "prompt": "<FRAME_NAME_PROMPT_2>", 
        "Count": "<NUMBER_OF_SUMMARIES_2>", 
        "Example IDs": "<URN_ID_2_1, URN_ID_2_2, URN_ID_2_3>"
    },
    {...}
  ]
}
\end{lstlisting}
\vspace{1em}

\textbf{LLM Answer:} 

[FRAME CATEGORIES ]
\vspace{1em}
\end{onehalfspacing}
\end{tcolorbox}

Lastly, all the outputted joint files are combined into one large set of generated frames while merging duplicates. This is then used to build the final list of selected frame classes to be used for frame classification of the formed sentences. 

\subsubsection{Frame Classification} \label{method_section_appendix_prompts_frames_i_likert}
This section details the final step of the framework, which is the actual classification of the frames, after selecting the frames and their respective commands. In the first step, we ask the LLM to summarize the sentence frame once more. Then, we use a chain-of-thought loop to assess how well each possible frame fits the sentence on a 7-point Likert scale using the frame summary from the first step to increase the quality of the fit assessment. After the LLM returns all the fit values, we filter for the highest fit values. The labels with the highest fit are then shown to the LLM again together with the sentence, and where we task the LLM to return the final frame label for the sentence or frame labels if the two are equally well fitting. 

\begin{tcolorbox}[colframe=black!75!white, colback=white, boxrule=0.5mm, width=\textwidth, sharp corners, left=1mm, right=1mm, top=1mm, bottom=1mm]
\begin{onehalfspacing}
\textbf{Step One:}

\textbf{System:} 
Read the sentence provided from a speech on artificial intelligence and summarize the frame present.

According to Robert Entman's definition where, framing suggests that frames select some aspects of a perceived reality and make them more salient in a communicating text, 
in such a way as to promote a particular problem definition, causal interpretation, moral evaluation, and/or treatment recommendation for the item described.

Summarize the identified frame in no more than two sentences.
\vspace{1em}

\textbf{User:} 

Please write a summary of the argumentative, indicative and or conceptual frame about AI present in the following sentence:

[SENTENCE]

Make sure the summary captures the core argument or perspective highlighted in the sentence and is no longer than two sentences.
\vspace{1em}

\textbf{LLM Answer:} 

[Frame Summary]
\vspace{1em}
\end{onehalfspacing}
\end{tcolorbox}

\begin{tcolorbox}[colframe=black!75!white, colback=white, boxrule=0.5mm, width=\textwidth, sharp corners, left=1mm, right=1mm, top=1mm, bottom=1mm]
\begin{onehalfspacing}
\textbf{Step Two:}

\textbf{System:} 

Read the sentence provided from a speech on artificial intelligence and summarize the frame present. 

According to Robert Entman's definition where framing suggests that frames select some aspects of a perceived reality and make them more salient in a communicating text, in such a way as to promote a particular problem definition, causal interpretation, moral evaluation, and/or treatment recommendation for the item described.

Determine if the frame could be classified as a [FRAME CATEGORY] frame.
\vspace{1em}

\textbf{User:} 

Sentence: [SENTENCE]

Please decide if the  [FRAME CATEGORY] frame is present in the sentence or not?
\vspace{1em}

\textbf{Assistant:}

[Frame Summary]
\vspace{1em}

\textbf{User:} 

CONTEXT:

I have the following SENTENCE:

[SENTENCE]

I also have a frame named [FRAME CATEGORY] with the following PROMPT:

[FRAME CATEGORY PROMPT]
\vspace{1em}

TASK:

How well does the above SENTENCE fit with the presented frame PROMPT? 

Formulate a one-sentence RATIONALE of your thought process and provide a resulting ANSWER of ONE of the following multiple-choice options, including just the NUMBER: 

- 1: Strongly Disagree

- 2: Disagree

- 3: Slightly Disagree

- 4: Neither Agree nor Disagree

- 5: Slightly Agree
    
- 6: Agree

- 7: Strongly Agree

Respond with ONLY the RATIONALE and the NUMBER in a VALID JSON Format structured like this: 
\{"Rationale": "<one-sentence rationale>", "Fit": "<Number>","Frame": "<Frame Name>"\}
\vspace{1em}

\textbf{LLM Answer:} 

[FRAME FIT JSON]
\vspace{1em}
\end{onehalfspacing}
\end{tcolorbox}

\begin{tcolorbox}[colframe=black!75!white, colback=white, boxrule=0.5mm, width=\textwidth, sharp corners, left=1mm, right=1mm, top=1mm, bottom=1mm]
\begin{onehalfspacing}
\textbf{Step Three:}

\textbf{System:} 

Read the sentence provided from a speech on artificial intelligence and select the most prominent and best-fitting frame from a given list of Frames.

According to Robert Entman's definition where framing suggests that frames select some aspects of a perceived reality and make them more salient in a communicating text, 
in such a way as to promote a particular problem definition, causal interpretation, moral evaluation, and/or treatment recommendation for the item described.

Determine which of the following frames fits best for the sentence (You can choose up to two frames): [FINAL FRAME CATEGORIES]
\vspace{1em}

\textbf{User:} 

Sentence: [SENTENCE]

Please decide which of the following frames is the most prominent one in the sentence: 

[FINAL FRAME CATEGORIES]
\vspace{1em}

\textbf{Assistant:}

Some Thoughts for the different Frames regarding the Sentence:

[FINAL FRAME RATIONALE SENTENCES]
\vspace{1em}

\textbf{User:} 

CONTEXT:

I have the following SENTENCE:

[SENTENCE]
\vspace{1em}

TASK:

Please decide which of the following FRAMES is the most prominent/fitting one in the sentence:

[FINAL FRAME CATEGORIES]

Respond with ONLY the FRAMES that fit best. 

You are allowed to return either ONE or TWO frames (preferably one) in the following format: <FRAME> OR <FRAME 1 | FRAME 2>
\vspace{1em}

\textbf{LLM Answer:} 

[FRAME LABEL]
\vspace{1em}
\end{onehalfspacing}
\end{tcolorbox}

\subsection{Prompts Frame Analysis II} \label{method_appendix_prompts_f2}
This outlines the framework used for frame generation and classification in the context of parliamentary speeches on artificial intelligence. The process uses an LLM in a sophisticated chain of functions to filter sentences with frames, summarize the frames, generate frame classes out of these summaries, and finally label the frames within the sentences after evaluating and selecting suggested frame classes by the author. 

\subsubsection{Frame Detection}
The first stage of the framework is identifying the frames in sentences. This involves a chain-of-thought interaction with the LLM, resulting in a binary classification that labels the presence or absence of a frame. 

\begin{tcolorbox}[colframe=black!75!white, colback=white, boxrule=0.5mm, width=\textwidth, sharp corners, left=1mm, right=1mm, top=1mm, bottom=1mm]
\begin{onehalfspacing}
\textbf{Interaction One:}

\textbf{System:} 

Determine whether the following paragraph from a newspaper about encryption exhibits a frame (framing effect) on/about encryption.

According to Robert Entman's definition where, framing suggests that frames select some aspects of a perceived reality and make them more salient in a communicating text 
in such a way as to promote a particular problem definition, causal interpretation, moral evaluation, and/or treatment recommendation for the item described.

Assess if specific language choices, focus points, or implied assumptions in the sentence promote a distinct perspective or argument, indicative of a frame.
\vspace{1em}

\textbf{User:} 

Article Title: [Title] 

Text: [Paragraph]

Is a frame present regarding encryption (yes, no)? Provide some thoughts.
\vspace{1em}

\textbf{LLM Answer:} 

[THOUGHTS]
\vspace{1em}
\end{onehalfspacing}
\end{tcolorbox}

\begin{tcolorbox}[colframe=black!75!white, colback=white, boxrule=0.5mm, width=\textwidth, sharp corners, left=1mm, right=1mm, top=1mm, bottom=1mm]
\begin{onehalfspacing}
\textbf{Interaction Two:}

\textbf{System:} 

Determine whether the following paragraph from a newspaper about encryption exhibits a frame (framing effect) on/about encryption.

According to Robert Entman's definition where, framing suggests that frames select some aspects of a perceived reality and make them more salient in a communicating text 
in such a way as to promote a particular problem definition, causal interpretation, moral evaluation, and/or treatment recommendation for the item described.

Assess if specific language choices, focus points, or implied assumptions in the sentence promote a distinct perspective or argument, indicative of a frame.
\vspace{1em}

\textbf{User:} 

Article Title: [Title] 

Text: [Paragraph] 

Is a frame present regarding encryption (yes, no)? Provide some thoughts.
\vspace{1em}

\textbf{Assistant:} 

[THOUGHTS]
\vspace{1em}

\textbf{User:} 

Now answer with just the correct category (yes, no). Please answer with just Yes or No.
\vspace{1em}

\textbf{LLM Answer:} 

[LABEL] (Yes, or No)
\vspace{1em}
\end{onehalfspacing}
\end{tcolorbox}

\subsubsection{Frame Summarizing} \label{methods_section_appendix_prompts_frame_ii_summary}
All sentences with a frame are then summarized with the LLM via Zero-Shot. This involves a system prompt and a user prompt, which guide the large language model to generate concise summaries of the frames found in a sentence. The system prompt includes a definition of a frame and an explanation of the task, instructing the LLM to summarize the sentence with a frame in no more than ten words.

\begin{tcolorbox}[colframe=black!75!white, colback=white, boxrule=0.5mm, width=\textwidth, sharp corners, left=1mm, right=1mm, top=1mm, bottom=1mm]
\begin{onehalfspacing}
\textbf{System:} 

Analyze the newspaper paragraph provided and identify any frame present.

According to Robert Entman's definition where framing suggests that frames select some aspects of a perceived reality and make them more salient in a communicating text, 
in such a way as to promote a particular problem definition, causal interpretation, moral evaluation, and/or treatment recommendation for the item described.

Assess if specific language choices, focus points, or implied assumptions in the sentence promote a distinct perspective or argument, indicative of a frame.

Summarize the identified frame in no more than 16 words.
\vspace{1em}

\textbf{User:} 

Please write a summary of the argumentative, indicative and or conceptual frame about encryption/cryptography present in the following newspaper paragraph in 8 to 16 words: [Praragraph]

Make sure to ensure the summary captures the core argument and perspective highlighted in the sentence.
\vspace{1em}

\textbf{LLM Answer:} 

[FRAME SUMMARY]
\vspace{1em}
\end{onehalfspacing}
\end{tcolorbox}

\subsubsection{Frame Generation}
To generate all possible frame classes, the frame generation process uses a series of zero-shot interactions together with the previously obtained frame summaries to generate frame class suggestions. This involves exhaustively drawing 100 summaries simultaneously and generating a maximum of 9 frame classes per call. This approach ensures a comprehensive list of possible frames, as each frame summary is seen at least once, with 20 \% of the 100 retained for the next call, while 80 new ones are added. Hence, the process begins by initializing the necessary variables and setting configurations for the LLM. Then, the loop iterates over the data frame containing frame summaries, selecting and sampling summaries to generate frame class suggestions. Thus, the system prompt includes the task and a definition of the framing concept, instructing the LLM to categorize the frame summaries into no more than 9 classes. The prompt provides the specific instructions and the summaries to the LLM, ensuring the response includes the frame class name and count of summaries in a machine-readable JSON format. 

\begin{tcolorbox}[colframe=black!75!white, colback=white, boxrule=0.5mm, width=\textwidth, sharp corners, left=1mm, right=1mm, top=1mm, bottom=1mm]
\begin{onehalfspacing}
\textbf{System:} 

Categorize the following set of frame summaries from a set of newspaper paragraphs with encryption frames present.
According to Robert Entman's definition where framing involves the selection and emphasis of aspects in a situation or issue to promote specific problem definitions, causal interpretations, moral evaluations, and treatment recommendations. 

This shapes the social reality and guides the audience's understanding.

Please come up with a set of categories but no more than 1 to 9 for these examples.
\vspace{1em}

\textbf{User:} 

I have these frame summaries:

[FRAME SUMMARIES]

Please come up with a maximum 9 frame classes and write a 2-4 word name for the frame category, the number of times this frame occurs in this sample and an associated chatGPT Prompt that can serve as an Instruction to determine if a new sentence contains this frame (framing effect) or not.

Please respond ONLY with a valid JSON in the following format (No yapping!):

\begin{lstlisting}
{
  "frame-categories": [
    {
        "frame": "<FRAME_NAME_1>", 
        "prompt": "<FRAME_NAME_PROMPT_1>", 
        "Count": "<NUMBER_OF_SUMMARIES_1>", 
        "Example IDs": "<URN_ID_1_1, URN_ID_1_2, URN_ID_1_3>"
    },
    {
        "frame": "<FRAME_NAME_2>", 
        "prompt": "<FRAME_NAME_PROMPT_2>", 
        "Count": "<NUMBER_OF_SUMMARIES_2>", 
        "Example IDs": "<URN_ID_2_1, URN_ID_2_2, URN_ID_2_3>"
    },
    {...}
  ]
}
\end{lstlisting}
\vspace{1em}

\textbf{LLM Answer:} 

[FRAME CATEGORIES ]
\vspace{1em}
\end{onehalfspacing}
\end{tcolorbox}

Lastly, all the outputted files are combined into one large set of generated frames while merging duplicates. This is then used to build the final list of selected frame classes to be used for frame classification of the formed sentences. 

\subsubsection{Frame Classification}
This section details the final step of the framework, which is the actual classification of the frames, after selecting the frames and their respective commands. In the first step, we ask the LLM to summarize the sentence frame once more. Then, we use a chain-of-thought loop to assess how well each possible frame fits the sentence on a 7-point Likert scale using the frame summary from the first step to increase the quality of the fit assessment. After the LLM returns all the fit values, we filter for the highest fit values. The labels with the highest fit are then shown to the LLM again together with the sentence, where we task the LLM to return the final frame label for the sentence or frame labels if the two are equally well fitting. 

\begin{tcolorbox}[colframe=black!75!white, colback=white, boxrule=0.5mm, width=\textwidth, sharp corners, left=1mm, right=1mm, top=1mm, bottom=1mm]
\begin{onehalfspacing}
\textbf{Step One:}

\textbf{System:} 
Read the paragraph provided from a news paper article and summarize the frame present.

According to Robert Entman's definition where framing suggests that frames select some aspects of a perceived reality and make them more salient in a communicating text, 
in such a way as to promote a particular problem definition, causal interpretation, moral evaluation, and/or treatment recommendation for the item described.

Summarize the identified frame in no more than two sentences.
\vspace{1em}

\textbf{User:} 

Please write a summary of the argumentative, indicative and or conceptual frame about encryption/cryptography present in the following paragraph:

[Paragraph]

Make sure the summary captures the core argument or perspective highlighted in the paragraph and is no longer than two sentences.
\vspace{1em}

\textbf{LLM Answer:} 

[Frame Summary]
\vspace{1em}
\end{onehalfspacing}
\end{tcolorbox}

\begin{tcolorbox}[colframe=black!75!white, colback=white, boxrule=0.5mm, width=\textwidth, sharp corners, left=1mm, right=1mm, top=1mm, bottom=1mm]
\begin{onehalfspacing}
\textbf{Step Two:}

\textbf{System:} 

Read the paragraph provided from a newspaper article and summarize the frame present. 

According to Robert Entman's definition where framing suggests that frames select some aspects of a perceived reality and make them more salient in a communicating text, 
in such a way as to promote a particular problem definition, causal interpretation, moral evaluation, and/or treatment recommendation for the item described.

Determine if the frame could be classified as a [FRAME CATEGORY] frame.
\vspace{1em}

\textbf{User:} 

Please write a summary of the argumentative, indicative and or conceptual frame about encryption/cryptography present in the following paragraph: 

[Paragraph]

Please decide if the  [FRAME CATEGORY] frame is present in the sentence or not?
\vspace{1em}

\textbf{Assistant:}

[Frame Summary]
\vspace{1em}

\textbf{User:} 

CONTEXT:

I have the following PARAGRAPH: [PARAGRAPH]

I also have a frame named [FRAME CATEGORY] with the following PROMPT:

[FRAME CATEGORY PROMPT]
\vspace{1em}

TASK:

How well does the above PARAGRAPH fit with the presented frame PROMPT? 

Formulate a one-sentence RATIONALE of your thought process and provide a resulting ANSWER of ONE of the following multiple-choice options, including just the NUMBER: 

- 1: Strongly Disagree

- 2: Disagree

- 3: Slightly Disagree

- 4: Neither Agree nor Disagree

- 5: Slightly Agree
    
- 6: Agree

- 7: Strongly Agree

Respond with ONLY the RATIONALE and the NUMBER in a VALID JSON Format structured like this: 
\{"Rationale": "<one-sentence rationale>", "Fit": "<Number>","Frame": "<Frame Name>"\}
\vspace{1em}

\textbf{LLM Answer:} 

[FRAME FIT JSON]
\vspace{1em}
\end{onehalfspacing}
\end{tcolorbox}

\begin{tcolorbox}[colframe=black!75!white, colback=white, boxrule=0.5mm, width=\textwidth, sharp corners, left=1mm, right=1mm, top=1mm, bottom=1mm]
\begin{onehalfspacing}
\textbf{Step Three:}

\textbf{System:} 

Read the paragraph provided from a news paper article on encryption / cryptography and select the most prominent and best-fitting frame from a given list of Frames.

According to Robert Entman's definition where framing suggests that frames select some aspects of a perceived reality and make them more salient in a communicating text, 
in such a way as to promote a particular problem definition, causal interpretation, moral evaluation, and/or treatment recommendation for the item described.

Determine which of the following frames fits best for the paragraph (You can choose up to two frames): [FINAL FRAME CATEGORIES]
\vspace{1em}

\textbf{User:} 

Paragraph: [SENTENCE]

Please decide which of the following FRAMES is the most prominent/fitting one in the paragraph:

[FINAL FRAME CATEGORIES]
\vspace{1em}

\textbf{Assistant:}

Some Thoughts for the different Frames regarding the sentence:

[FINAL FRAME RATIONALE SENTENCES]
\vspace{1em}

\textbf{User:} 

CONTEXT:

I have the following PARAGRAPH:

[PARAGRAPH]
\vspace{1em}

TASK:

Please decide which of the following FRAMES is the most prominent/fitting one in the paragraph:

[FINAL FRAME CATEGORIES]

Respond with ONLY the FRAMES that fit best. 

You are allowed to return either ONE or TWO frames (preferably one) in the following format: <FRAME> OR <FRAME 1 | FRAME 2>
\vspace{1em}

\textbf{LLM Answer:} 

[FRAME LABEL]
\vspace{1em}
\end{onehalfspacing}
\end{tcolorbox}

\subsection{Prompts Topic Modeling} \label{method_appendix_prompts_t1}
\subsubsection{Topic Summarizing} \label{methods_section_appendix_prompts_topic_i_summary}
All articles are first summarized with the LLM via Zero-Shot. This involves a system prompt and a user prompt, which guide the LLM to generate concise summaries of the main topic found in the article. The system prompt includes a definition of a topic and an explanation of the task, instructing the LLM to summarize the sentence with a frame in no more than ten words.

\begin{tcolorbox}[colframe=black!75!white, colback=white, boxrule=0.5mm, width=\textwidth, sharp corners, left=1mm, right=1mm, top=1mm, bottom=1mm]
\begin{onehalfspacing}
\textbf{System:} 

Analyze the text provided and identify the topic.

For this task, a 'topic' is understood as the main subject or theme discussed in an article, encapsulating the core ideas and issues being addressed. Each topic corresponds to a specific category that best describes the content of the article.

Summarize the identified topic in no more than two sentences.
\vspace{1em}

\textbf{User:} 

Please write a summary of the main subject or theme discussed in the following article in one to two  sentences: [ARTICLE]

Make sure to ensure the summary captures the core subject or theme discussed in a detailed manner.
\vspace{1em}

\textbf{LLM Answer:} 

[FRAME SUMMARY]
\vspace{1em}
\end{onehalfspacing}
\end{tcolorbox}

\subsubsection{Frame Generation}
To generate all possible topic categories, the topic generation process uses a series of zero-shot interactions alongside previously obtained topic summaries to produce category suggestions. This involves simultaneously selecting 100 summaries and generating up to 21 topic categories per iteration. By doing so, each topic summary is encountered at least once, with 20 \% of the current set retained for continuity, and 80 new summaries introduced in each subsequent iteration.

The procedure starts with initializing the necessary variables and configuring the LLM. The interaction then proceeds over a data frame of topic summaries, systematically sampling these summaries to generate topic category proposals. To guide the model, the system prompt outlines the task and includes a definition of the 'topic' concept, instructing the LLM to group the topic summaries into no more than 21 categories. The prompt includes explanatory instructions and choice summaries, ensuring that the LLM response provides category names along with the count of summaries for each category in machine-readable JSON format.

\begin{tcolorbox}[colframe=black!75!white, colback=white, boxrule=0.5mm, width=\textwidth, sharp corners, left=1mm, right=1mm, top=1mm, bottom=1mm]
\begin{onehalfspacing}
\textbf{System:} 

Categorize the following set of topic summaries from a set of news articles.
For this task, a 'topic' is understood as the main subject or theme discussed in an article, encapsulating the core ideas and issues being addressed. Each topic corresponds to a specific category that best describes the content of the article.

Please come up with a set of topic categories but no more than 10 to 21 for these examples. Ensure the categories are comprehensive enough to cover closely related subtopics but specific enough to maintain clear and distinct themes/topics. This means topics should be identified based on themes that can encompass related subtopics without being overly general or narrowly specific.
\vspace{1em}

\textbf{User:} 

I have these topic summaries:

[TOPIC SUMMARIES]

Please come up with a maximum 21 topic categories and write a 1 to 4 word title for the topic category,  the total count of this topic in this sample and an associated chatGPT PROMPT that can serve as an Instruction to determine if a new article is about this topic or not. Always add a MISCELLANEOUS class for uncategorizable topics.

Please respond ONLY with a VALID JSON in the following format (No yapping!):

\begin{lstlisting}
{
  "frame-categories": [
    {
        "topic": "<TOPIC_NAME_1>", 
        "prompt": "<TOPIC_NAME_PROMPT_1>", 
        "Count": "<NUMBER_OF_SUMMARIES_1>", 
        "Example IDs": "<URN_ID_1_1, URN_ID_1_2, URN_ID_1_3>"
    },
    {
        "topic": "<TOPIC_NAME_2>", 
        "prompt": "<TOPIC_NAME_PROMPT_2>", 
        "Count": "<NUMBER_OF_SUMMARIES_2>", 
        "Example IDs": "<URN_ID_2_1, URN_ID_2_2, URN_ID_2_3>"
    },
    {...}
  ]
}
\end{lstlisting}
\vspace{1em}

\textbf{LLM Answer:} 

[FRAME CATEGORIES ]
\vspace{1em}
\end{onehalfspacing}
\end{tcolorbox}

Lastly, all the outputted joint files are combined into one large set of generated topics while merging duplicates. This is then used to build the final list of selected topic classes to be used for the topic classification of the articles.

\subsubsection{Frame Classification}
This section details the final step of the framework, which is the actual classification of the topics, after selecting the topics and their respective commands. In the first step, we ask the LLM to summarize the article once more. Then, we use a chain-of-thought loop to assess how well each possible topic fits the article on a 7-point Likert scale using the topic summary from the first step to increase the quality of the fit assessment. After the LLM returns all the fit values, we filter for the highest fit values. The labels with the highest fit are then shown to the LLM again together with the article, and where we task the LLM to return the final topic label for the article or frame labels if the two are equally well fitting. 

\begin{tcolorbox}[colframe=black!75!white, colback=white, boxrule=0.5mm, width=\textwidth, sharp corners, left=1mm, right=1mm, top=1mm, bottom=1mm]
\begin{onehalfspacing}
\textbf{Step One:}

\textbf{System:} 
Analyze the text provided and identify the topic.

For this task, a 'topic' is understood as the main subject or theme discussed in an article, encapsulating the core ideas and issues being addressed. Each topic corresponds to a specific category that best describes the content of the article.

Summarize the identified topic in no more than two sentences.

\vspace{1em}

\textbf{User:} 

Please write a summary of the main subject or theme discussed in the following article in one to two  sentences: 

[ARTICLE]

Make sure to ensure the summary captures the core subject, topic or theme discussed in a detailed manner and is no longer than two sentences.
\vspace{1em}

\textbf{LLM Answer:} 

[Frame Summary]
\vspace{1em}
\end{onehalfspacing}
\end{tcolorbox}

\begin{tcolorbox}[colframe=black!75!white, colback=white, boxrule=0.5mm, width=\textwidth, sharp corners, left=1mm, right=1mm, top=1mm, bottom=1mm]
\begin{onehalfspacing}
\textbf{Step Two:}

\textbf{System:} 

Analyze the text provided and identify the topic.

For this task, a 'topic' is understood as the main subject or theme discussed in an article, encapsulating the core ideas and issues being addressed. Each topic corresponds to a specific category that best describes the content of the article.

Determine if the topic could be classified with the following topic name: [TOPIC CATEGORY]

\vspace{1em}

\textbf{User:} 

Article: [ARTICLE]

Please decide if the  [TOPIC CATEGORY] TOPIC is present in the ARTICLE or not?
\vspace{1em}

\textbf{Assistant:}

[Frame Summary]
\vspace{1em}

\textbf{User:} 

CONTEXT:

I have the following ARTICLE:

[ARTICLE]

I also have a TOPIC named [TOPIC CATEGORY] with the following PROMPT:

[TOPIC CATEGORY PROMPT]
\vspace{1em}

TASK:

How well does the above ARTICLE fit with the presented topic PROMPT? 

Formulate a one-sentence RATIONALE of your thought process and provide a resulting ANSWER of ONE of the following multiple-choice options, including just the NUMBER: 

- 1: Strongly Disagree

- 2: Disagree

- 3: Slightly Disagree

- 4: Neither Agree nor Disagree

- 5: Slightly Agree
    
- 6: Agree

- 7: Strongly Agree

Respond with ONLY the RATIONALE and the NUMBER in a VALID JSON Format structured like this: 
\{"Rationale": "<one-sentence rationale>", "Fit": "<Number>","Topic": "<Topic Name>"\}
\vspace{1em}

\textbf{LLM Answer:} 

[FRAME FIT JSON]
\vspace{1em}
\end{onehalfspacing}
\end{tcolorbox}

\begin{tcolorbox}[colframe=black!75!white, colback=white, boxrule=0.5mm, width=\textwidth, sharp corners, left=1mm, right=1mm, top=1mm, bottom=1mm]
\begin{onehalfspacing}
\textbf{Step Three:}

\textbf{System:} 

Read the article provided from a news paper and select the most prominent and best-fitting topic from a given list of possible topics.

For this task, a 'topic' is understood as the main subject or theme discussed in an article, encapsulating the core ideas and issues being addressed. Each topic corresponds to a specific category that best describes the content of the article.

Determine if the topic could be classified with the following topic name (You can choose up to two topics): [FINAL TOPIC CATEGORIES]
\vspace{1em}

\textbf{User:} 

Article: [ARTICLE]

Please decide which of the following TOPICS is the most prominent/fitting one in the article:

[FINAL TOPIC CATEGORIES]
\vspace{1em}

\textbf{Assistant:}

Some Thoughts for the different Topics regarding the Article:

[FINAL TOPIC RATIONALE SENTENCES]
\vspace{1em}

\textbf{User:} 

CONTEXT:

I have the following ARTICLE:

[ARTICLE]
\vspace{1em}

TASK:

Please decide which of the following TOPICS is the most prominent/fitting one in the article:

[FINAL TOPIC CATEGORIES]

Respond with ONLY the TOPIC that fit best. 

You are allowed to return either ONE or TWO topics (preferably one) in the following format: <TOPIC> OR <TOPIC 1 | TOPIC 2>
\vspace{1em}

\textbf{LLM Answer:} 

[TOPIC LABEL]
\vspace{1em}
\end{onehalfspacing}
\end{tcolorbox}

\end{document}